\documentclass[times,twocolumn,final]{elsarticle}

\usepackage{framed,multirow}

\usepackage{amssymb}
\usepackage{latexsym}

\usepackage{multirow}
\usepackage{graphicx}
\usepackage{booktabs}
\usepackage{caption}
\usepackage{array}
\usepackage{makecell}

\usepackage{url}
\usepackage{xcolor}

\usepackage{pifont}

\usepackage[hidelinks]{hyperref}
\definecolor{newcolor}{rgb}{.8,.349,.1}

\usepackage[ruled,vlined,algo2e]{algorithm2e}
\SetKwComment{Comment}{$\triangleright$\ }{}

\usepackage{mathtools}
\usepackage{amsmath}

\begin{document}


\begin{frontmatter}

\title{AMA-SAM: Adversarial Multi-Domain Alignment of Segment Anything Model for High-Fidelity Histology Nuclei Segmentation}

\author[1]{Jiahe Qian}
\author[1]{Yaoyu Fang}
\author[1]{Jinkui Hao}
\author[1]{Bo Zhou\corref{cor1}}
\cortext[cor1]{Corresponding author.}
\ead{bo.zhou@northwestern.edu}

\address[1]{Department of Radiology, Northwestern University, Chicago, IL, USA}





\begin{abstract}
Accurate segmentation of cell nuclei in histopathology images is essential for numerous biomedical research and clinical applications. However, existing cell nucleus segmentation methods only consider a single dataset (i.e., primary domain), while neglecting to leverage supplementary data from diverse sources (i.e., auxiliary domains) to reduce overfitting and enhance the performance. Although incorporating multiple datasets could alleviate overfitting, it often exacerbates performance drops caused by domain shifts. In this work, we introduce Adversarial Multi-domain Alignment of Segment Anything Model (AMA-SAM) that extends the Segment Anything Model (SAM) to overcome these obstacles through two key innovations. First, we propose a Conditional Gradient Reversal Layer (CGRL), a multi-domain alignment module that harmonizes features from diverse domains to promote domain-invariant representation learning while preserving crucial discriminative features for the primary dataset. Second, we address SAM’s inherent low-resolution output by designing a High-Resolution Decoder (HR-Decoder), which directly produces fine-grained segmentation maps in order to capture intricate nuclei boundaries in high-resolution histology images. To the best of our knowledge, this is the first attempt to adapt SAM for multi-dataset learning with application to histology nuclei segmentation. We validate our method on several publicly available datasets, demonstrating consistent and significant improvements over state-of-the-art approaches. 

\end{abstract}


\end{frontmatter}


\section{Introduction}
Nuclei segmentation is a critical task in digital pathology analysis, playing a central role in disease diagnosis, histological research, and personalized medicine. In tissue sections and microscopic images, cell nuclei's morphology, size, and spatial distribution provide vital information for assessing tissue health, identifying pathological alterations, and predicting disease progression. 

Although deep learning and computer vision have enabled the development of numerous neural network-based approaches for histology nuclei segmentation [\cite{naylor2018segmentation, zhou2019cia, zhou2019irnet, graham2019hover, Micro-Net, koohbanani2020nuclick, chen2020blendmask, gong2021style, he2021cdnet, chen2023cpp, nam2023pronet, he2023toposeg}], existing methods are trained on a single dataset, rendering them prone to overfitting to the specific characteristics of that dataset. Moreover, a naive fusion of multiple datasets often leads to further performance degradation because differences in imaging acquisition systems, staining protocols, and tissue types across datasets introduce substantial domain shifts, which can confound the learning process when data are simply mixed. Consequently, developing a robust histology nuclei segmentation model that can effectively leverage multi-source data to enhance performance on the target dataset (i.e. primary dataset/domain), while efficiently leveraging supplementary data from diverse sources (i.e. auxiliary domains) remains a significant challenge, with no universal solution established to date.

The Segment Anything Model (SAM) [\cite{kirillov2023segment}] represents a significant advancement in universal image segmentation, owing to its robust architecture and large-scale pretraining that enable remarkable adaptability and accuracy across a broad spectrum of natural images. Despite these advances, SAM exhibits some limitations in histological nuclei segmentation. First, SAM employs a fixed output resolution of $256 \times 256$ pixels, necessitating upsampling of segmentation outputs via interpolation to match the original image's higher resolution required for detailed histology analysis. This process can blur nuclei boundaries and adversely affect downstream analytical tasks, such as the accurate computation of area or perimeter especially for small nuclei. Secondly, while SAM can be fine-tuned on diverse datasets, the persistent challenge of domain discrepancy arising from differences in image characteristics and annotation standards continues to compromise its performance when training on multiple datasets. These limitations underscore the need for enhanced methodologies that can effectively handle high-resolution inputs and reconcile domain differences to achieve robust segmentation outcomes.

To address these challenges, we propose an innovative SAM-based framework, called Adversarial Multi-Domain Alignment of Segment Anything Model (AMA-SAM), which leverages collaborative training on a primary dataset and auxiliary datasets from diverse sources to achieve high-fidelity high-resolution histological nuclei segmentation on the primary dataset. Specifically, our contributions include: \textbf{(1) Multi-domain Alignment via Conditional Gradient Reversal Layer (CGRL)}: We introduce a novel CGRL for SAM to facilitate effective adversarial multi-domain alignment. By selectively applying gradient reversal, the CGRL aligns features from auxiliary domains to the primary domain while preserving the integrity of the primary domain’s features. This targeted alignment enables the integration of complementary information from auxiliary datasets, thereby significantly enhancing segmentation performance on the primary dataset. \textbf{(2) High-Resolution Decoder (HR-Decoder)}: To address SAM’s low output resolution, we design an HR-Decoder that generates high-resolution segmentation results. By freezing SAM's original decoder and incorporating additional slice tokens, along with our Multi-token Slices Producer and Pixel Ensemble Module, our method reconstructs detailed high-resolution segmentation, preserving fine segmentation boundaries and structural details. \textbf{(3) Comprehensive Validation}: We rigorously evaluate AMA-SAM across a variety of datasets and experimental scenarios. The results demonstrate the framework’s superior segmentation performance on the primary dataset, achieved by effectively integrating complementary information from auxiliary datasets. Our proposed framework combines innovative domain alignment strategies with architectural enhancements to overcome key limitations of SAM in high-resolution histological nuclei segmentation. 

\section{Related Work}

\noindent \textbf{Nuclei Segmentation:} Early methods for nuclei segmentation predominantly relied on handcrafted features and classical image processing techniques [\cite{yang2006nuclei, ali2012integrated, liao2016automatic}]. With the advent of deep learning, a multitude of convolutional neural network-based approaches have been developed for histology nuclei segmentation [\cite{naylor2018segmentation, zhou2019cia, zhou2019irnet, graham2019hover, Micro-Net, koohbanani2020nuclick, chen2020blendmask, gong2021style, he2021cdnet, chen2023cpp, nam2023pronet, he2023toposeg, schmidt2018stardist, chen2024edge, vuola2019mask, hollandi2022nucleus, chen2024sam}]. For instance, HoVer-Net [\cite{graham2019hover}] addresses the problem of overlapping nuclei by predicting horizontal and vertical displacement maps, whereas Micro-Net [\cite{Micro-Net}] enhances the classical U-Net architecture [\cite{ronneberger2015u}] with multi-scale feature extraction to capture fine morphological details. Semi-automatic methods such as NuClick [\cite{koohbanani2020nuclick}] incorporate user-provided scribbles for guidance, and more recent approaches including CDNet [\cite{he2021cdnet}], CPP-Net [\cite{chen2023cpp}], and PROnet [\cite{nam2023pronet}] further improve boundary delineation by employing directional feature maps, complementary boundary masks, or offset maps. More recently, UN-SAM [\cite{chen2024sam}] fine-tuned SAM using an automatic learning prompt strategy, achieving state-of-the-art performance in nuclei segmentation on single dataset.

\noindent \textbf{Multi-dataset training:} Multi-dataset training [\cite{shim2023multidataset, wu2023pointprompt, zhang2021cover, likhosherstov2021polyvit, liu2021multi}] has emerged as a promising strategy to enhance model generalizability and robustness by leveraging data from heterogeneous sources. Prior studies have demonstrated that integrating diverse datasets can mitigate dataset bias [\cite{hinami2018multimodal}] and boost performance in tasks such as object detection [\cite{zhou2022simple, wang2019towards, wang2021multidomain, chen2021openvocab, zhang2023mmd3d, soumfontez2023mdt3d, chen2023scaledet, wang2024uni2det}] and semantic segmentation [\cite{bevandic2019simultaneous, ghadiyaram2019large, liang2022multi}]. However, existing methods of multi-dataset co-training exhibit specific limitations that render them suboptimal for histology nuclei segmentation tasks. First, most existing methods for multi-dataset co-training in segmentation primarily address label inconsistencies across datasets [\cite{bevandic2019simultaneous, ros2016training, lambert2020mseg}], rather than resolving domain gaps of images. Second, existing multi-dataset training approaches typically treat all datasets equally to achieve balanced improvements. In practice, significant variability in data characteristics leads to a dilution of performance on the primary dataset compared to training exclusively on it. Third, inconsistent annotation standards and dataset-specific noise further exacerbate these challenges, undermining model stability and accuracy.

\section{Methods}

\begin{figure*}
    \centering
    \includegraphics[width=1\linewidth]{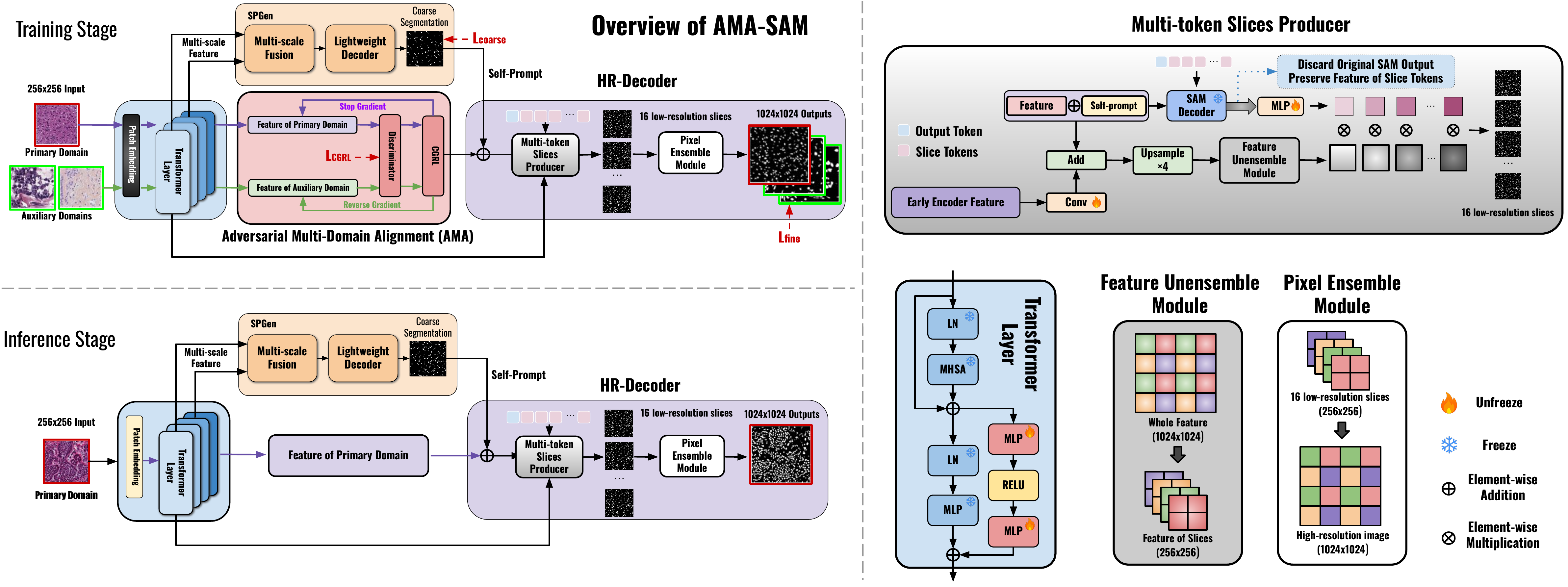}
    \caption{Flowchart for training and inference of AMA-SAM and its crucial components. Our AMA-SAM is the first model which can utilize auxiliary datasets with distinct domain distributions to enhance the nuclei segmentation performance on the primary dataset.}
    \label{fig:3}
\end{figure*}

Our model includes three key components: AMA-SAM overall design (Section \ref{method:pipeline}), CGRL (Section \ref{method:cgrl}), and HR-Decoder (Section \ref{method:hrdecoder}) in the AMA-SAM. We will provide a detailed explanation of the model's training process (Section \ref{method:objective}) and implementation details (Section \ref{method:implementation}). Datasets, baselines, and evaluation metrics will be introduced in Section \ref{method:datasets}. The overall process is illustrated in Figure \ref{fig:3}. 

\subsection{Overall Pipeline of AMA-SAM} \label{method:pipeline}

Figure \ref{fig:3} illustrates the training and inference process of our AMA-SAM. During training, the pretrained Segment Anything Model (SAM) serves as the foundational base network. Multi-source datasets from diverse domains are fed into the encoder, where we integrate a module at each transformer layer that consists of two MLP layers and a ReLU activation function (as indicated by the red and yellow boxes in the bottom left of Figure~\ref{fig:3}) to adapt the feature representations. To align the feature distributions between the primary and auxiliary datasets while minimizing the impact on the primary domain, we introduce the Conditional Gradient Reversal Layer (CGRL) (Section \ref{method:cgrl}) for multi-domain alignment. Additionally, an automatic prompt generation module called SPGen (same in UN-SAM [\cite{chen2024sam}]) is employed, where aligned features are processed by a lightweight decoder that produces coarse segmentation maps. These maps then act as segmentation prompts, guiding the fine segmentation process without the need for manual intervention. Finally, we design a High-Resolution Decoder (Section \ref{method:hrdecoder}) to directly generate full-resolution nuclei segmentation outputs, ensuring precise delineation of boundaries and fine structural details. Throughout the training process, all parameters in the pretrained SAM model are frozen, and only the newly incorporated modules are optimized.

During inference, the framework is streamlined to operate solely on the primary domain. The input image is processed through the model without CGRL and other domain alignment components to directly produce high-quality, high-resolution nuclei segmentation results.

\subsection{Conditional Gradient Reversal Layer} \label{method:cgrl}

Conditional Gradient Reversal Layer (CGRL) is a multi-domain alignment strategy designed to preserve and enhance performance on the primary dataset while leveraging diverse auxiliary datasets to further boost segmentation accuracy on the primary domain. The layer and associated loss designs are detailed as follows.

\noindent\textbf{Conditional Gradient Reversal Layer (CGRL):} To mitigate undesired degradation on the primary dataset, CGRL refines the gradient reversal process by selectively blocking the reversed gradients for samples originating from the primary dataset. In the forward pass, CGRL operates identically to a standard GRL:
\begin{equation}
\label{eq:cgrl_forward}
{X}'_i = \mathrm{CGRL}({X}_i),
\end{equation}
where ${X}_i$ represents the input feature of the $i$th sample, and ${X}'_i$ denotes the feature after passing through the CGRL. This operation does not alter the feature values during the forward pass.

During the backward pass, CGRL differentiates between primary and auxiliary samples by conditionally reversing gradients:
\begin{equation}
\label{eq:cgrl_backward}
\frac{\partial \mathcal{L}}{\partial {X}_i} = 
\begin{cases}
-\lambda\, \frac{\partial \mathcal{L}}{\partial {X}'_i}, & \text{if sample $i$ is from an auxiliary dataset},\\
0, & \text{if sample $i$ is from the primary dataset}.
\end{cases}
\end{equation}

Here, $\mathcal{L}$ denotes the total loss, and $\lambda$ is a scalar factor controlling the strength of gradient reversal. By setting the gradient to zero for primary dataset samples, CGRL prevents any adversarial influence from altering the feature representations of the primary domain. Conversely, for auxiliary samples, the gradients are reversed and scaled by $\lambda$, encouraging the model to learn domain-invariant features that are indistinguishable between primary and auxiliary domains.

\noindent\textbf{Domain Discriminator and CGRL Loss:} To facilitate effective feature alignment, we incorporate a domain discriminator $D(\cdot)$ that predicts the origin of each sample, distinguishing between the primary and auxiliary datasets. For a batch containing $N$ samples, let $y_i \in \{0,1\}$ indicate the dataset origin of the $i$th sample, where $y_i = 1$ if the sample is from the primary dataset and $y_i = 0$ otherwise. The discriminator outputs $D({X}'_i)$, the probability that the $i$th sample originates from the primary domain.

To address potential class imbalances due to varying dataset sizes, we introduce weighting factors $w_{\text{main}}$ and $w_{\text{aux}}$ for primary and auxiliary samples, respectively. The total adversarial loss for training with CGRL is formulated as:

{\scriptsize
\begin{equation}
\label{eq:adv}
\mathcal{L}_{\text{CGRL}} 
= -\frac{1}{N} \sum_{i=1}^{N} 
\Bigl[\, w_{\text{main}}\, y_i \log D({X}'_i) 
\;+\; w_{\text{aux}}\, (1 - y_i) \log \bigl(1 - D({X}'_i)\bigr) \Bigr].
\end{equation}
}

where $y_i$ represents whether the $i$th sample originates from the main dataset, $D({X}'_i)$ is the domain discriminator's predicted probability that the sample comes from the main domain, and $w_{\text{main}}$ and $w_{\text{aux}}$ are the weights for the main dataset and auxiliary datasets, respectively. These weights are set inversely proportional to the number of images in each dataset, thereby compensating for any class imbalances of the domains.

The integration of CGRL within the encoder allows the model to leverage auxiliary datasets effectively without compromising the performance on the primary dataset. By only reversing gradients for auxiliary samples, CGRL ensures that the primary dataset's feature space remains stable and well-preserved, while auxiliary datasets contribute complementary information that refines the feature representation and enhances segmentation performance on the primary domain.

\subsection{High-Resolution Decoder} \label{method:hrdecoder}

As shown in Figure~\ref{fig:3}, High-Resolution Decoder (HR-Decoder) is designed to address the resolution limitations inherent in SAM, particularly for high-resolution nuclei segmentation in histology images. To retain the benefits of SAM’s large-scale pretraining and avoid instability caused by excessive parameter adjustments, we freeze the parameters of the original SAM decoder. In addition, we introduce 16 new slice tokens into the HR-Decoder, each corresponding to a distinct region of the input image. These tokens are integrated with the original output and prompt tokens and processed through two distinct sub-modules: the \textbf{Multi-token Slices Producer} and the \textbf{Pixel Ensemble Module}.

The \textbf{Multi-token Slices Producer} consists of two sub-branches, as illustrated in the up right of Figure~\ref{fig:3}. In the first sub-branch, the newly introduced 16 slice tokens interact with all other tokens, including the original SAM output tokens, via a self-attention mechanism. This interaction allows each slice token to capture both global and local features by attending to the full set of tokens, enabling the model to leverage both image-to-token and token-to-image attention for enhanced feature integration.

In the second sub-branch, the encoder's output features $F_{\text{encoder}}$, initially of size $64 \times 64$, are first processed through a convolutional layer which increases its resolution to size $256 \times 256$. Then, the upsampled features are combined with $F_{\text{mask}}$. After combination, the resulting feature map is further upsampled by a factor of 4 to $1024 \times 1024$ and then processed through the \textbf{Feature Unensemble Module} (gray box in the bottom right of Figure~\ref{fig:3}), which splits the enriched feature map into 16 distinct slice features, each corresponding to one of the 16 slice tokens. The complete process can be expressed as:

\begin{equation}
\{ F_{\text{slice}_i} \}_{i=1}^{16} = \operatorname{FUM}\left(\operatorname{UpSample}\left(\operatorname{Conv}(F_{\text{encoder}}) + F_{\text{mask}}\right)\right)
\end{equation}

where $\text{Conv}$ denotes the convolution operation, $\text{UpSample}$ refers to the upsampling step, and the $\operatorname{FUM}$ represents the Feature Unensemble Module, which divides the upsampled feature map into 16 individual slice features $F_{\text{slice}_i}$, each corresponding to a specific slice token. 

The two sub-branches are then integrated. In this stage, we apply an MLP to update the feature of slice tokens $T'_{\text{slice}_i}$ and perform pointwise multiplication with the corresponding slice features $F_{\text{slice}_i}$, producing 16 low-resolution slices. This process is expressed as:

\begin{equation}
S_{\text{slice}_i} = \text{MLP}(T'_{\text{slice}_i}) \times F_{\text{slice}_i}
\end{equation}

where $\text{MLP}(T'_{\text{slice}_i})$ represents the updated $i^{th}$ Slice Token, and $S_{\text{slice}_i}$ is the resulting low-resolution slice output.

Then, the \textbf{Pixel Ensemble Module} (white box in the bottom right of Figure~\ref{fig:3}) reassembles these 16 low-resolution slices into a single high-resolution segmentation output. Initially, the 16 single-channel slices (each $256 \times 256$) are stacked along the channel dimension to form a tensor of shape $(B, 16, 256, 256)$. Then, by redistributing the channel dimension into the spatial dimensions using a scaling factor of four, the tensor is transformed into a unified single-channel image of shape $(B, 1, 1024, 1024)$. This two-stage process effectively converts multiple low-resolution slices into a high-resolution segmentation output, ensuring enhanced resolution and edge detail. It should be noted that $1024 \times 1024$ is just the default output resolution of the HR-Decoder. Sometimes the resolution of the nuclei segmentation image itself is lower than $1024 \times 1024$. Therefore, after obtaining the segmentation results, it is necessary to downsample according to the actual size of the image to ensure that the final segmentation result matches the original image resolution.

\subsection{Objective Functions} \label{method:objective}
The pretrained SAM is utilized to initialize the network parameters, including those of the encoder and decoder. These parameters, having undergone extensive training on large-scale datasets, exhibit robust feature extraction and initial segmentation capabilities. To retain the benefits of these pretrained parameters, we freeze all original SAM parameters during the training process and fine-tune only the newly added network layers. The objective during training is to minimize the segmentation loss, adversarial loss, and automatic prompt generation loss simultaneously. The total loss function is defined as follows:

\begin{equation} \mathcal{L} = \mathcal{L}_{\text{fine}} + \alpha \mathcal{L}_{\text{CGRL}} + \beta \mathcal{L}_{\text{coarse}}, \end{equation}

\noindent where $\mathcal{L}_{\text{fine}}$ denotes the segmentation loss, which measures the discrepancy between the model’s predicted segmentation and the true labels. $\mathcal{L}_{\text{CGRL}}$ represents the CGRL adversarial training loss, aimed at aligning feature distributions across different datasets to mitigate the impact of domain variations. $\mathcal{L}_{\text{coarse}}$ corresponds to the automatic prompt generation loss, ensuring that the prompts generated by the SPGen module effectively guide the segmentation process. The parameters $\alpha$ and $\beta$ (both set to 1) serve as weighting factors to balance the contributions of each loss component within the overall loss function.

\subsection{Implementation Details} \label{method:implementation}
All experiments were conducted on a single NVIDIA 4090 GPU using the PyTorch framework. To ensure the fairness of the experiments, we adopted the same training settings and configurations for implementing all cell nucleus segmentation methods. Specifically, all SAM models used the ViT-B [\cite{dosovitskiy2020image}] architecture as the image encoder to ensure comparability between different methods.

In terms of optimization, we chose the Adam optimizer to optimize the model parameters. The training process lasted for 30 epochs. For the setting of the learning rate, we used a strategy inversely proportional to the number of images, with a dynamic adjustment using an exponential decay strategy with a decay factor of 0.98. For example, when training with 40 images, we set the initial learning rate to $2 \times 10^{-4}$.

\subsection{Datasets, Baselines, and Evaluation Metrics} \label{method:datasets}
To comprehensively evaluate the performance of our method in cell nucleus segmentation tasks, we selected 4 representative and challenging public datasets: MoNuSeg [\cite{kumar2017dataset}], TNBC [\cite{naylor2018segmentation}], CryoNuSeg [\cite{mahbod2021cryonuseg}], and cpm17 [\cite{vu2019methods}]. These datasets cover various tissue types, imaging techniques, and annotation standards, providing a robust test of the model's adaptability and robustness in diverse scenarios.

\noindent \textbf{MoNuSeg} is a widely-used multi-organ cell nucleus segmentation benchmark, consisting of 44 high-resolution images from various organs such as lungs, liver, and kidneys, each with a resolution of 1000x1000 pixels. 

\noindent \textbf{TNBC} focuses on triple-negative breast cancer tissues and includes 50 high-resolution immunohistochemically stained images, each with a resolution of 512x512 pixels. 

\noindent \textbf{CryoNuSeg} is tailored for cell nucleus segmentation tasks in cryosectioned tissues, containing 30 images of various tissue types, encompassing different staining methods and imaging conditions, each with a resolution of 512x512 pixels. 

\noindent \textbf{cpm17} includes 64 images from different tissue origins and staining methods, with varying resolutions that enhance the diversity and complexity of the dataset.

To validate the effectiveness of our method, we conduct experiments on both nuclei semantic segmentation and instance segmentation tasks, selecting a diverse array of baseline methods for comparative analysis. For both semantic and instance segmentation tasks, U-Net [\cite{ronneberger2015u}], nnU-Net [\cite{isensee2021nnu}], and UN-SAM [\cite{chen2024sam}] serve as the baselines. Unet++ [\cite{isensee2021nnu}], ResUNet++ [\cite{jha2019resunet++}], and DoubleU-Net [\cite{jha2020doubleu}] are specifically employed as baselines for semantic segmentation. For instance segmentation exclusively, Mask-RCNN [\cite{he2017mask}], HoVer-Net [\cite{graham2019hover}], and CPP-Net [\cite{chen2023cpp}] are utilized. For semantic segmentation tasks, we utilized four metrics: Dice score (DSC), mean Intersection over Union (mIoU), F1 score, and Hausdorff Distance (HD). The DSC quantifies the overlap between predicted and actual regions. mIoU calculates the average overlap of segmentation results across all categories. F1 score offers a holistic assessment of the model's segmentation accuracy. HD evaluates the maximal discrepancy between predicted and actual boundaries. In instance segmentation tasks, we adopted the Aggregated Jaccard Index (AJI), Detection Quality (DQ), Segmentation Quality (SQ), and Panoptic Quality (PQ) as metrics. AJI evaluates the overlap quality between detection results and real instances. DQ gauges the accuracy of instance detection. SQ addresses the precision of segmentation in detected instances. PQ, amalgamating DQ and SQ, provides a comprehensive measure of overall instance segmentation quality.

\section{Experimental Results}

\begin{figure*}
    \centering
    \includegraphics[width=1\linewidth]{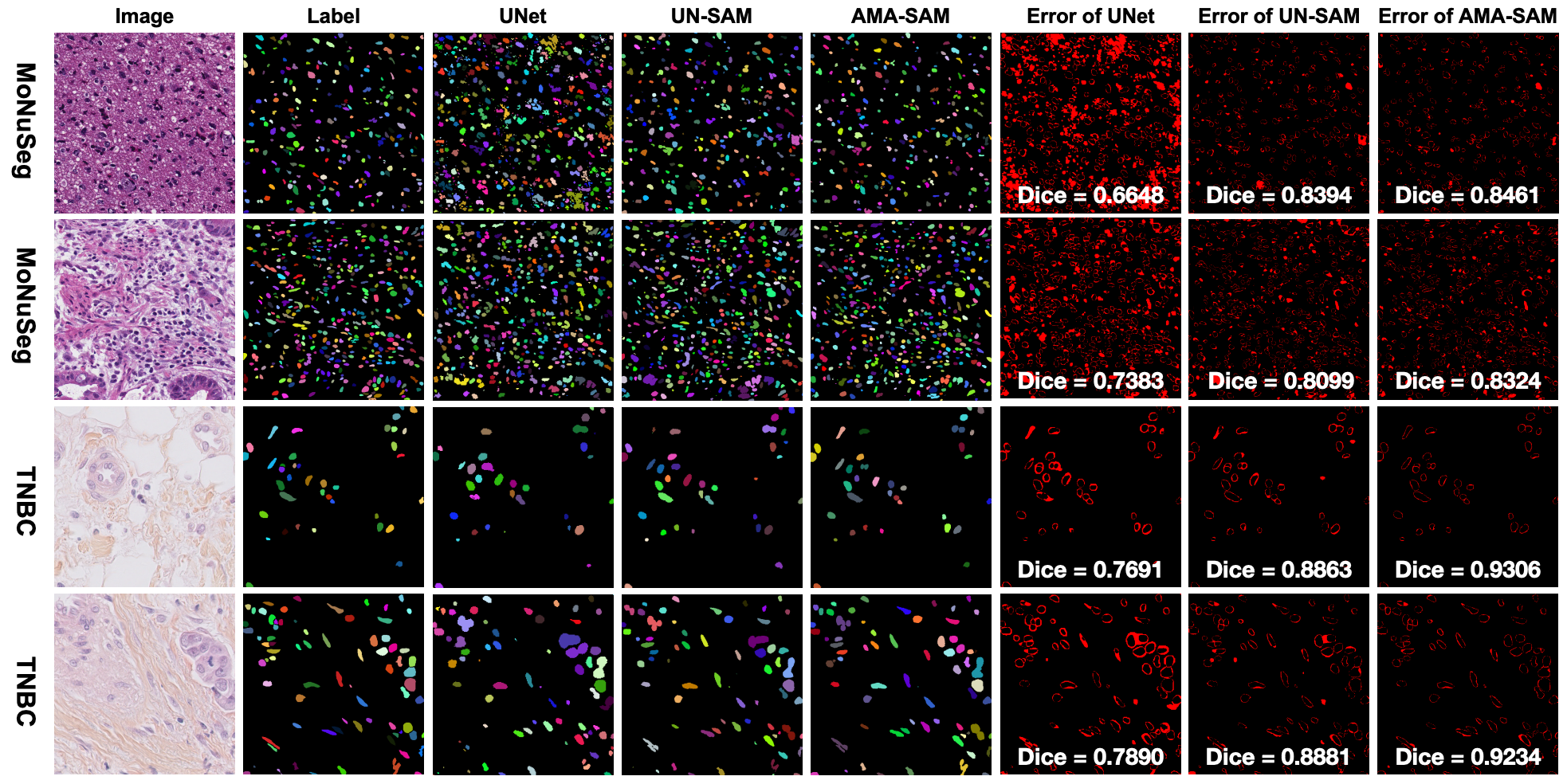}
    \caption{Visualization comparison of nuclei segmentation results for U-Net (3rd column), UN-SAM (4th column), and AMA-SAM (5th column) on MoNuSeg and TNBC datasets. The corresponding segmentation errors are reported on the right with the Dice score computed between the predictions and the human label (2nd column).}
    \label{fig:C1}
\end{figure*}

\begin{table*}[htbp]
    \centering
    \fontsize{8}{10}\selectfont
    \setlength{\tabcolsep}{3.5pt}
    \caption{Quantitative comparison of \textit{Semantic Segmentation} and \textit{Instance Segmentation} results on MoNuSeg and TNBC datasets. The upper half contains the results of semantic segmentation evaluations, and the lower half contains the results of instance segmentation evaluations. M represents MoNuSeg, T represents TNBC, C represents CryoNuSeg, and CP represents cpm17. When training AMA-SAM, the evaluation dataset is used as the primary dataset during training, while the other datasets used in training serve as auxiliary datasets.}
    \label{tab:1}
    \begin{tabular}{c|ccccc|ccccc}
    \toprule
    \textbf{Semantic Segmentation} & \multicolumn{10}{c}{\textbf{Evaluation Dataset}} \\
    \cmidrule(lr){1-11}
    \multirow{2.5}{*}{Method} & \multicolumn{5}{c|}{\textbf{M}} & \multicolumn{5}{c}{\textbf{T}} \\
    \cmidrule(lr){2-11}
     & Training Dataset & DSC(\%) & mIoU & F1 & HD 
     & Training Dataset & DSC(\%) & mIoU & F1 & HD \\
    \midrule
    \multirow{2}{*}{U-Net} 
      & M & 74.06 & 60.25 & 75.57 & 18.94 & T & 80.64 & 67.62 & 81.00 & 38.39 \\
      & M + T + C + CP & 73.50 & 59.66 & 75.09 & 19.20 & T + M + C + CP & 79.76 & 66.45 & 80.28 & 39.73 \\
    \midrule
    \multirow{2}{*}{Unet++} 
      & M & 76.78 & 62.97 & 78.31 & 18.53 & T & 81.19 & 68.44 & 81.60 & 37.70 \\
      & M + T + C + CP & 75.77 & 61.63 & 77.43 & 18.78 & T + M + C + CP & 80.69 & 67.73 & 80.96 & 38.72 \\
    \midrule
    \multirow{2}{*}{ResUNet++} 
      & M & 77.96 & 64.05 & 78.85 & 19.45 & T & 76.81 & 62.55 & 77.53 & 33.20 \\
      & M + T + C + CP & 77.48 & 63.50 & 69.65 & 19.53 & T + M + C + CP & 76.12 & 61.60 & 76.75 & 34.47 \\
    \midrule
    \multirow{2}{*}{DoubleU-Net} 
      & M & 78.20 & 64.43 & 78.81 & 19.01 & T & 81.59 & 68.98 & 81.83 & 40.66 \\
      & M + T + C + CP & 77.05 & 63.09 & 78.00 & 19.25 & T + M + C + CP & 81.05 & 68.41 & 81.40 & 41.70 \\
    \midrule
    \multirow{2}{*}{nnU-Net} 
      & M & 81.09 & 68.28 & 81.24 & 18.15 & T & 82.11 & 69.74 & 82.35 & 31.71 \\
      & M + T + C + CP & 80.67 & 67.63 & 80.88 & 18.33 & T + M + C + CP & 81.67 & 69.23 & 82.01 & 32.48 \\
    \midrule
    \multirow{2}{*}{UN-SAM} 
      & M & 84.17 & 72.93 & 84.27 & 16.34 & T & 85.72 & 75.02 & 85.81 & 26.83 \\
      & M + T + C + CP & 83.13 & 72.02 & 83.33 & 16.51 & T + M + C + CP & 84.63 & 73.67 & 84.75 & 28.30 \\
    \midrule
    \multirow{2}{*}{AMA-SAM} 
      & M\textsuperscript{\dag} & 84.51 & 73.52 & 84.69 & 16.27 & T\textsuperscript{\dag} & 86.36 & 75.77 & 86.44 & 26.02 \\
      & M\textsuperscript{\dag} + T + C + CP & \textbf{85.12$^*$} & \textbf{74.24$^*$} & \textbf{85.21$^*$} & \textbf{16.13$^*$} 
          & T\textsuperscript{\dag} + M + C + CP & \textbf{87.10$^*$} & \textbf{76.76$^*$} & \textbf{87.10$^*$} & \textbf{24.76$^*$} \\
    \toprule
    \textbf{Instance Segmentation} & \multicolumn{10}{c}{\textbf{Evaluation Dataset}} \\
    \cmidrule(lr){1-11}
    \multirow{2.5}{*}{Method} & \multicolumn{5}{c|}{\textbf{M}} & \multicolumn{5}{c}{\textbf{T}} \\
    \cmidrule(lr){2-11}
      & Training Dataset & AJI & DQ & SQ & PQ
      & Training Dataset & AJI & DQ & SQ & PQ \\
    \midrule
    \multirow{2}{*}{U-Net}
      & M & 34.98 & 50.94 & 68.59 & 34.96 & T & 50.52 & 68.02 & 74.24 & 50.52 \\
      & M + T + C + CP & 33.76 & 49.77 & 68.28 & 33.55 & T + M + C + CP & 48.43 & 65.75 & 73.51 & 48.55 \\
    \midrule
    \multirow{2}{*}{nnU-Net}
      & M & 45.22 & 65.12 & 71.78 & 48.81 & T & 56.48 & 75.77 & 77.97 & 59.30 \\
      & M + T + C + CP & 43.37 & 63.23 & 71.13 & 47.02 & T + M + C + CP & 54.07 & 73.47 & 76.33 & 56.67 \\
    \midrule
    \multirow{2}{*}{Mask-RCNN}
      & M & 36.15 & 52.17 & 70.03 & 37.46 & T & 51.04 & 67.33 & 75.01 & 50.83 \\
      & M + T + C + CP & 34.78 & 51.00 & 69.55 & 36.02 & T + M + C + CP & 49.08 & 65.51 & 74.04 & 48.61 \\
    \midrule
    \multirow{2}{*}{HoVer-Net}
      & M & 46.76 & 66.37 & 70.92 & 47.58 & T & 56.45 & 76.19 & 77.65 & 59.11 \\
      & M + T + C + CP & 46.05 & 65.51 & 70.70 & 46.90 & T + M + C + CP & 54.75 & 75.32 & 76.70 & 57.43 \\
    \midrule
    \multirow{2}{*}{CPP-Net}
      & M & 44.63 & 60.57 & 70.53 & 42.44 & T & 52.20 & 69.17 & 74.75 & 51.67 \\
      & M + T + C + CP & 43.37 & 59.03 & 70.11 & 40.88 & T + M + C + CP & 50.73 & 67.76 & 74.11 & 50.22 \\
    \midrule
    \multirow{2}{*}{UN-SAM}
      & M & 50.59 & 70.37 & 72.88 & 50.86 & T & 59.48 & 78.13 & 78.64 & 61.47 \\
      & M + T + C + CP & 49.13 & 68.89 & 72.57 & 49.60 & T + M + C + CP & 58.34 & 77.00 & 78.02 & 60.41 \\
    \midrule
    \multirow{2}{*}{AMA-SAM}
  & M\textsuperscript{\dag} & 51.23 & 70.90 & 73.03 & 51.42 
  & T\textsuperscript{\dag} & 59.76 & 78.49 & 78.77 & 61.76 \\
  & M\textsuperscript{\dag} + T + C + CP & \textbf{52.24$^*$} & \textbf{71.71$^*$} 
    & \textbf{73.31$^*$} & \textbf{52.36$^*$}
  & T\textsuperscript{\dag} + M + C + CP & \textbf{60.53$^*$} & \textbf{79.24$^*$} 
    & \textbf{79.20$^*$} & \textbf{62.35$^*$} \\
\bottomrule
\end{tabular}

\vspace{1ex}
\footnotesize{$^\dag$ indicates the primary dataset when training AMA-SAM. $^*$ means the difference between AMA-SAM (multi-dataset version) and UN-SAM (single-dataset version) is significant at $p<0.05$.}

\end{table*}

\subsection{Main Results}
A qualitative comparison of segmentation results from different methods across various datasets is illustrated in Figure~\ref{fig:C1}. In the first histology example, taken from a liver tissue sample in MoNuSeg, UNet trained on all available datasets (i.e., MoNuSeg+TNBC+CryoNuSeg+cpm17) exhibits significant segmentation errors. Compared to human annotation, we observe a high number of false-positive segmentations in non-nuclei regions. Additionally, while true-positive nuclei regions are detected, their segmentation boundaries are poorly defined, often appearing enlarged. This results in a low DSC of 0.6648. Fine-tuning SAM for the MoNuSeg dataset using UN-SAM leads to notable improvements; however, we observe a loss in precision for smaller nuclei, likely due to the low-resolution output from SAM. In contrast, our AMA-SAM method further refines segmentation, achieving the highest DSC of 0.8461 for this sample. A similar pattern is observed in the second example, which is a lung tissue sample from MoNuSeg. For breast cancer samples from TNBC, where nuclei are more sparsely distributed, UNet achieves reasonable performance without significant false positives, possibly due to the sample’s cleaner background. However, nuclei are still prone to both over-segmentation and under-segmentation. While UN-SAM substantially improves upon UNet due to its large-scale pretraining, our AMA-SAM consistently delivers superior segmentation results, providing more precise boundary delineation and reducing mis-segmentation errors.

The quantitative comparison is reported in Table~\ref{tab:1} with comprehensive evaluation both in terms of nuclei semantic segmentation and instance segmentation. To ensure fair comparisons, all models are trained and tested under identical conditions.

\noindent\textbf{Semantic Segmentation} results are reported in the upper section of Table~\ref{tab:1} based on metrics including DSC, mIoU, F1, and HD on the MoNuSeg (M) and TNBC (T) datasets. When MoNuSeg is used as the primary dataset, conventional methods (i.e., those not based on SAM) achieve their best performance with nnU-Net obtaining a DSC of 81.09\% under single-dataset training. In contrast, UN-SAM, a SAM-based method, reaches a DSC of 84.17\% under the same condition. Notably, when auxiliary datasets (TNBC, CryoNuSeg, and cpm17) are incorporated during training, conventional methods tend to exhibit a decline in performance. For example, nnU-Net’s DSC decreases from 81.09\% to 80.67\% and UN-SAM’s from 84.17\% to 83.13\%. Conversely, AMA-SAM benefits from the integration of auxiliary data, with its DSC improving from 84.51\% to 85.12\% on MoNuSeg and from 86.36\% to 87.10\% on TNBC. These results indicate that although naive multi-dataset training can undermine the performance of conventional segmentation models, our targeted multi-domain alignment strategy in AMA-SAM effectively leverages auxiliary data to enhance segmentation accuracy on the primary dataset.

\noindent\textbf{Instance Segmentation} results are presented in the lower half of Table~\ref{tab:1}, which reports evaluation metrics including AJI, DQ, SQ, and PQ for both the MoNuSeg (M) and TNBC (T) datasets. Among conventional non-SAM methods, nnU-Net achieves an AJI of 45.22 on MoNuSeg under single-dataset training, while Mask-RCNN and HoVer-Net record AJI values of 36.15 and 46.76, respectively. In comparison, the SAM-based method UN-SAM attains an AJI of 50.59 on MoNuSeg with single-dataset training, though this decreases to 49.13 when auxiliary datasets are incorporated. In contrast, our proposed AMA-SAM achieves an AJI of 51.23 under single-dataset training on MoNuSeg, which further increases to 52.24 with the inclusion of auxiliary data. A similar trend is observed on the TNBC dataset, where UN-SAM's AJI decreases from 59.48 to 58.34 following the incorporation of auxiliary data, while AMA-SAM shows an overall performance improvement, highlighting its capacity to benefit from a multi-dataset training process.

Overall, our results show that AMA-SAM enhances segmentation accuracy on the primary dataset by effectively integrating multi-source data and mitigating domain discrepancies, establishing it as a robust solution for precise nuclei segmentation.

\subsection{Ablation Studies}

\noindent\textbf{Impact of Multi-domain Alignment:} To investigate the impact of our multi-domain alignment strategy on segmentation performance, we conduct a series of controlled experiments on the MoNuSeg dataset. Table~\ref{tab:3} summarizes the results under four distinct configurations: (1) training solely on the primary dataset without any auxiliary data; (2) incorporating auxiliary datasets (TNBC, CryoNuSeg, and cpm17) without employing multi-domain alignment; (3) integrating auxiliary datasets using a standard Gradient Reversal Layer (GRL) \cite{ganin2015unsupervised} for multi-domain alignment; and (4) utilizing our proposed Conditional Gradient Reversal Layer (CGRL) for multi-domain alignment.

When trained exclusively on the primary dataset, the model achieved a baseline DSC of 84.51\%. In contrast, naively adding auxiliary data without any alignment led to a slight performance decline, underscoring the detrimental effect of unaddressed domain discrepancies. When a standard GRL was employed, performance metrics showed modest improvement. However, our CGRL approach further improved the results, as evidenced by a higher DSC, enhanced mIoU and F1, and a reduced HD.

These findings clearly demonstrate that our targeted CGRL-based alignment not only mitigates the adverse effects of incorporating auxiliary datasets but also enhances the preservation and refinement of the primary dataset's feature representation, ultimately leading to improved segmentation accuracy.

\begin{table}[htbp]
\centering
\fontsize{8}{10}\selectfont
\setlength{\tabcolsep}{4.5pt}
\caption{Performance of Different Multi-domain Alignment Strategies on MoNuSeg with and without Auxiliary Data}
\label{tab:3}
\begin{tabular}{cccccc}
\toprule
\textbf{Aux. Datasets} & \textbf{Adv. Training} & \textbf{DSC(\%)} & \textbf{mIoU} & \textbf{F1} & \textbf{HD} \\
\midrule
 & - & 84.51 & 73.52 & 84.69 & 16.27 \\
\checkmark & - & 83.77 & 72.08 & 83.88 & 16.42 \\
\checkmark & GRL & 84.63 & 73.71 & 84.65 & 16.23 \\
\checkmark & CGRL & \textbf{85.12} & \textbf{74.24} & \textbf{85.21} & \textbf{16.13} \\
\bottomrule
\end{tabular}
\end{table}

\noindent\textbf{Impact of HR-Decoder:} We assess the efficacy of our proposed High-Resolution Decoder by comparing it with the original SAM decoder [\cite{kirillov2023segment}]. In these experiments, MoNuSeg is used as the primary dataset, while TNBC, CryoNuSeg, and cpm17 are used as auxiliary datasets. To ensure that any performance variations are solely due to differences in the decoder design, all other network components, including the encoder and training protocols, are kept constant.

Table~\ref{tab:4} summarizes the results on the MoNuSeg dataset. For instance, the SAM Decoder achieved a DSC of 84.62\%, while our HR-Decoder further elevated the DSC to 85.12\%, accompanied by improvements in mIoU, F1, and a reduction in HD. Such enhancements are indicative of more precise segmentation and superior boundary delineation.

These performance gains can be attributed to the HR-Decoder's design, which introduces 16 specialized slice tokens to capture localized image details. Overall, the experimental outcomes underscore the advantage of our HR-Decoder in extracting high-resolution details, validating its superiority over the previous decoders.

\begin{table}[htbp]
\centering
\caption{Effectiveness of Decoders on MoNuSeg Dataset}
\label{tab:4}
\begin{tabular}{ccccc}
\toprule
\textbf{Decoder} & \textbf{DSC(\%)} & \textbf{mIoU} & \textbf{F1} & \textbf{HD} \\
\midrule
SAM Decoder & 84.62 & 73.60 & 84.68 & 16.25 \\
\textbf{HR-Decoder} & \textbf{85.12} & \textbf{74.24} & \textbf{85.21} & \textbf{16.13} \\
\bottomrule
\end{tabular}
\end{table}

In Table \ref{tab:5}, we further examine the impact of the number of auxiliary datasets on segmentation performance using MoNuSeg as the primary dataset. In this experiment, the model is initially trained exclusively on MoNuSeg, establishing a baseline with a DSC of 84.51\%, mIoU of 73.52, F1 of 84.69, and a HD of 16.27. Subsequently, the TNBC dataset is incorporated as an auxiliary source, resulting in an improved DSC of 84.73\% and a corresponding enhancement in other metrics. When the CryoNuSeg dataset is further added, a slight additional gain is observed, and the integration of all three auxiliary datasets (TNBC, CryoNuSeg, and cpm17) yields the highest performance, with a DSC reaching 85.12\%, mIoU of 74.24, F1 of 85.21, and HD reduced to 16.13. 

These incremental improvements, ndicate that each auxiliary dataset contributes complementary information that refines the feature representation of the primary dataset. The progressive enhancement in performance metrics demonstrates the effectiveness of our multi-domain alignment strategy in mitigating domain discrepancies and harnessing diverse data sources, thereby substantially boosting segmentation accuracy on the primary dataset.

\begin{table}[htbp]
    \centering
    \fontsize{8}{10}\selectfont
    \setlength{\tabcolsep}{4.5pt}
    \caption{Experimental results when using different numbers of auxiliary datasets with MoNuSeg as the primary dataset.}
    \label{tab:5}
    \begin{tabular}{ccc|cccc}
    \toprule
    \multicolumn{3}{c|}{Auxiliary Dataset}
        & \multirow{2.5}{*}{DSC(\%)} 
        & \multirow{2.5}{*}{mIoU}
        & \multirow{2.5}{*}{F1} 
        & \multirow{2.5}{*}{HD} \\
    \cmidrule(lr){1-3}
    TNBC & CryoNuSeg & cpm17 
        & 
        & 
        & 
        & \\
    \midrule
    & & & 84.51 & 73.52 & 84.69 & 16.27 \\
    \checkmark & & & 84.73 & 73.74 & 84.76 & 16.22 \\
    \checkmark & \checkmark & & 84.88 & 74.11 & 84.92 & 16.18 \\
    \checkmark & \checkmark & \checkmark & \textbf{85.12} & \textbf{74.24} & \textbf{85.21} & \textbf{16.13} \\
    \bottomrule
    \end{tabular}
\end{table}

\section{Conclusion}

This paper presents a novel multi-dataset training framework, called AMA-SAM, for histology nucleus segmentation that effectively addresses the challenges of high-resolution imaging and multi-domain data integration. Our approach is built upon two key innovations. First, we introduce the Conditional Gradient Reversal Layer (CGRL), which aligns the feature distributions of auxiliary datasets with those of the primary dataset. This targeted alignment mitigates domain discrepancies while preserving the intrinsic characteristics of the primary data, thereby enabling the model to leverage diverse training information to improve its performance. Second, we develop a High-Resolution Decoder (HR-Decoder) that augments the original Segment Anything Model (SAM) by freezing its decoder and integrating 16 additional slice tokens. These tokens are processed through a Multi-token Slices Producer and a Pixel Ensemble Module to directly produce segmentation outputs at a resolution of 1024×1024, effectively eliminating upsampling artifacts and enhancing the clarity of nuclear boundaries. Extensive experiments and ablation studies confirm that our framework significantly improves segmentation accuracy, particularly in preserving fine details and nuclear morphology across multiple datasets, thus establishing it as a robust solution for high-resolution nucleus segmentation in biomedical applications.




\bibliographystyle{model2-names.bst}\biboptions{authoryear}
\bibliography{refs}

\end{document}